\documentclass[11pt,a4paper]{article}

\usepackage[utf8]{inputenc}
\usepackage[T1]{fontenc}
\usepackage{lmodern}
\usepackage[a4paper,margin=2.25cm]{geometry}
\usepackage{graphicx}
\usepackage{subcaption}
\usepackage{booktabs}
\usepackage{array}
\usepackage{amsmath,amssymb,amstext,amsthm}
\usepackage{microtype}
\usepackage{algorithm2e}
\usepackage[bottom]{footmisc}
\usepackage{url}
\usepackage{authblk}
\usepackage[hidelinks,bookmarks=false]{hyperref}
\usepackage{apalike}

\newcommand{\abcllm}{\textsc{ABC-LLM}}
\newcommand{\modelname}{Llama 3.1 8B Instruct}
\newcommand{\op}[1]{\texttt{#1}}

\title{From One-Shot Generation to Incremental Music Composition:\\
Adapting a General-Purpose Instruction LLM for Persistent Symbolic Editing}

\author[1]{André Ricardo Ducca Fernandes}
\author[2,1,3]{Jean-Pierre Briot}
\author[1]{Simone Diniz Junqueira Barbosa}
\author[1]{Hélio Côrtes Vieira Lopes}

\affil[1]{Department of Informatics (DI), Pontifícia Universidade Católica do Rio de Janeiro (PUC-Rio), Rio de Janeiro, RJ 22451-900, Brazil}
\affil[2]{Sorbonne Université, CNRS, LIP6, F-75005 Paris, France}
\affil[3]{CBAE, Universidade Federal do Rio de Janeiro (UFRJ), Rio de Janeiro, RJ 22250-020, Brazil}

\date{Preprint, September 2026}

\begin{document}
\maketitle

\begin{abstract}
Most music-generation systems are still framed and evaluated primarily as producers of complete outputs, whereas composition often proceeds through successive revisions to a shared musical artifact. This paper studies a different use of a general-purpose instruction-following large language model: not as a one-shot music generator, but as a reusable operator over an evolving symbolic score. We formulate incremental composition as a sequence of operation-aware state transitions over persistent ABC notation, with explicit requirements on what each operation may change and what it must preserve. The interaction includes two artifact-initialization variants and three editing operations---chord addition, inpainting, and transposition. We instantiate the formulation by adapting Llama 3.1 8B Instruct with Low-Rank Adaptation (LoRA) on 496,038 operation-aware dialogue records derived from Irish traditional music. The comparison with the unadapted model is used to test the feasibility of learning this interaction contract, not to claim novelty for fine-tuning itself. Across 500 dialogues per model (1,750 attempted output states), checker admission rises from 29.37\% to 99.37\%, while compliance conditional on admission rises from 0.7205 to 0.9798. Strict eligibility for reference-relative musical-feature analysis increases from 14 to 1,548 outputs, and Longest Common Subsequence analysis does not show a systematic increase in high-overlap sequences relative to held-out baselines under the specified protocol. The results support the technical feasibility of persistent, operation-aware symbolic editing with a general-purpose instruction LLM. They do not establish superior musical quality or human--AI co-creativity, which remain questions for musician-centered evaluation.
\end{abstract}

\vspace{0.4em}

\section{\uppercase{Introduction}}
\label{sec:introduction}

Music composition is not only a problem of producing complete pieces from scratch. Composers commonly revise existing material: they preserve some decisions, replace others, move between global and local changes, and continue from the current state of a piece. A useful analogy is text editing. A writing assistant that could only regenerate an entire document whenever one paragraph should change would provide generation, but only limited support for writing as an iterative activity. Music composition raises the same distinction: a musician may wish to modify a passage, add harmony, transpose material, or preserve most of a score while changing only a specified part.

This interaction-centered view is supported by human-centered studies of AI-assisted music creation. In case studies of music creators, transparency and user control emerged as central concerns; the authors argue that AI creation tools should support continued iteration while making clear how users can edit, modify, and manipulate AI-produced material around each iteration \cite{newman2023humanai}. Recent systems also make revision itself increasingly explicit: BeatEdit recasts symbolic generation as editing a draft rather than synthesizing from scratch, while MusiChat maintains an active composition state across conversational turns \cite{gu2026beatedit,liao2026musichat}.

This paper investigates a complementary computational question: \emph{can a general-purpose instruction-following LLM be specialized to participate reliably in an incremental symbolic-composition process in which successive natural-language operations transform a persistent musical artifact while preserving operation-specific invariants?} The aim is deliberately not to design a dedicated music-editing architecture and then demonstrate that it can edit music. Instead, we test how far operation semantics, state continuity, and preservation behavior can be learned through instruction supervision by a general-purpose model. We provisionally call the adapted model \abcllm.

The interaction uses five canonical operations: \op{Create a melody}, \op{Create a melody with chords}, \op{Add chords}, \op{Inpainting}, and \op{Transpose}. The first two initialize or replace the artifact; the others transform its current state. Current scope is global for \op{Add chords} and \op{Transpose} and selected bars for \op{Inpainting}. Each operation specifies what may change and what must be preserved, and each output becomes the artifact for the next turn. Transposition can, of course, be implemented deterministically without an LLM; here it tests whether the same instruction-following mechanism can execute heterogeneous operations under one persistent interaction contract.

ABC is used pragmatically rather than as a contribution in itself. Its text form is directly compatible with language-model input/output while remaining compact, inspectable, and suitable for lead-sheet-like melody with chord symbols. The broader interaction formulation is not conceptually tied to ABC.

Our empirical study adapts \modelname\ with LoRA on simulated multi-turn dialogues. Evaluation keeps distinct four questions that are easy to conflate: whether the response contains usable ABC, whether the requested operation and its preservation invariants are satisfied, how selected musical characteristics compare with reference music, and how much generated material overlaps with training material. The base/adapted comparison is therefore a feasibility experiment for the proposed interaction contract, not a contribution based merely on the observation that task-specific fine-tuning improves task performance.

The paper makes three contributions. First, it formulates conversational symbolic composition as operation-specific state transitions over a persistent artifact, explicitly separating interaction continuity, transformation scope, and preservation invariants. Second, it operationalizes this formulation through a large operation-aware dialogue dataset and parameter-efficient adaptation of a general-purpose instruction LLM. Third, it presents a denominator-aware evaluation that separates operation compliance, reference-relative musical characteristics, and corpus-relative memorization evidence, and uses it to test whether the learned interaction contract is reliable under the stated domain and protocol. An experimental NONOTO integration demonstrates how this model behavior can be embedded in a score-oriented workflow.

The remainder of the paper reviews and positions related systems, presents the method and evaluation protocol, reports the empirical comparison, and discusses the prototype, scope of the claims, limitations, and future research directions.

\section{\uppercase{Related Work}}
\label{sec:related}

\begin{figure*}[!t]
\centering
\includegraphics[width=0.96\textwidth]{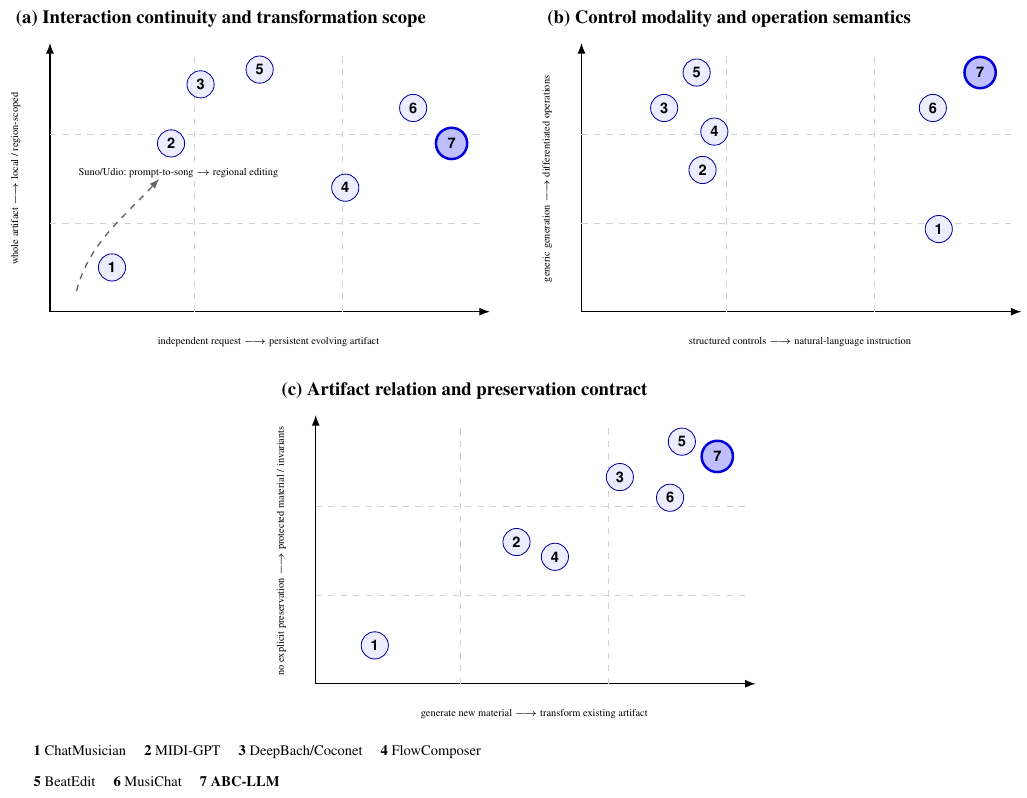}
\caption{Three qualitative projections of the focused design space used to position related systems: (a) interaction continuity versus transformation scope, (b) control modality versus operation semantics, and (c) relation to the existing artifact versus preservation contract. Marker positions are explanatory, not metric scores or rankings. The dashed Suno/Udio arrow is a contextual product-design trend, not an academic comparison.}
\label{fig:design-space}
\end{figure*}

Computational music systems differ along several independent dimensions. A single categorization such as ``structured controls'', ``natural language'', or ``conversation'' can therefore hide important distinctions: local rewriting does not imply persistent interaction, natural-language control does not imply editing, and a persistent workflow does not determine the underlying model architecture. Likewise, \emph{from scratch}, \emph{one shot}, \emph{global}, and \emph{end-to-end} describe different properties rather than synonyms.
Figure~\ref{fig:design-space} previews three qualitative projections of the focused design space used below; Section~2.3 returns to the dimensions and research gap after reviewing the closest systems.

\subsection{Symbolic Revision and Assisted Composition}
Symbolic representations make notes, chords, meter, and other musical structure explicitly manipulable, and have long supported local revision. DeepBach uses pseudo-Gibbs sampling to regenerate selected positions under constraints, while Coconet learns masked reconstruction and supports nonlinear rewriting by blocked Gibbs sampling \cite{hadjeres2017deepbach,huang2017coconet}. These works are important precedents: local editing and inpainting are not new claims of the present paper.

FlowComposer provides a different precedent: assisted lead-sheet composition. Its statistical and constraint-based engine is embedded in a workflow where the composer retains control and can work from a blank or partial score \cite{papadopoulos2016flowcomposer}. Composer's Assistant supports interactive multitrack MIDI infilling while preserving surrounding context \cite{malandro2023composer}. MIDI-GPT similarly targets computer-assisted multitrack composition with a dedicated Transformer that supports generation, continuation, track- and bar-level infilling, and structured controls for instrumentation, style, density, polyphony, and duration \cite{pasquier2025midigpt}. NONOTO complements these systems at the interface level as a model-agnostic score editor for interactive composition by inpainting \cite{bazin2019nonoto}.

BeatEdit makes the editing perspective explicit. It introduces a dedicated symbolic representation and typed edit mechanisms for error correction, accompaniment refinement, and segment completion, emphasizing that much musical creation consists of locating what should change while preserving what should not \cite{gu2026beatedit}. This is close to our preservation concern but differs computationally: BeatEdit engineers an edit-specific architecture, whereas our question is whether a general-purpose instruction LLM can learn the interaction contract directly from operation-aware supervision.

\subsection{Natural Language and Persistent Conversation}
LLMs make textual symbolic music especially interesting because user instructions and musical notation can share one model interface. ChatMusician adapts LLaMA2 through continual pre-training and fine-tuning on ABC, treating music as a second language and demonstrating text-conditioned understanding and generation \cite{yuan2024chatmusician}. Our question is narrower: after returning a melody, a later \op{Add chords} request must preserve that exact melody, and a subsequent \op{Inpainting} request must act on the harmonized state rather than on a newly generated sample.

MusiChat is a particularly close contemporary system. It explicitly targets the limitations of prompt-and-regenerate workflows and maintains an active composition state across natural-language interactions, but it does so with a hybrid architecture that combines LLM-based reasoning and intent routing with a symbolic music engine \cite{liao2026musichat}. BeatEdit and MusiChat appeared nearly simultaneously in July 2026 \cite{gu2026beatedit,liao2026musichat}. We interpret this timing not as a priority contest but as independent evidence that editable and iterative music generation is becoming a distinct research direction.

\subsection{A Focused Design Space and Research Gap}
A broader design-space perspective has recently been developed for Music Co-Creation Systems. Silva's systematic mapping of 46 systems organizes co-creative design decisions around musical tasks, exchanged inputs, temporal granularity, interaction protocol, interaction paradigms, generation method, and stylistic divergence \cite{silva2026designspace}. Our purpose is narrower: to isolate properties especially relevant to incremental transformation of a persistent musical artifact.

We therefore distinguish nine analytical lenses: temporal organization; state persistence; relation to an existing artifact; transformation scope; operation semantics; preservation contract; control modality; musical representation; and computational organization. These are not scores or a hierarchy. Figure~\ref{fig:design-space} shows three qualitative projections that are particularly relevant here. A system can be close to \abcllm\ in one projection and different in another. The dashed Suno/Udio trajectory in Figure~\ref{fig:design-space}(a) is contextual rather than a scientific baseline: both commercial systems now expose region-level replacement/editing in addition to prompt-based generation \cite{suno2026editor,udio2026sessions}.

The specific gap addressed here lies at the intersection of four properties. First, the user communicates differentiated compositional operations in ordinary language. Second, the returned symbolic score is the authoritative state for the next turn. Third, editing operations define observable preservation invariants, so correctness concerns both what changes and what must remain stable. Fourth, these behaviors are learned by a general-purpose instruction LLM through state-transition supervision rather than supplied by a dedicated editing architecture or symbolic transformation engine at inference time.

This gap is deliberately scoped. We do not claim that local editing, inpainting, assisted composition, natural-language music generation, or persistent interaction are individually new, nor that fine-tuning an LLM is itself a scientific contribution. The question is whether a general-purpose instruction model can learn the resulting \emph{interaction contract} reliably enough to support incremental symbolic composition, and how that capability should be evaluated without confusing syntax, operation compliance, musical characterization, and training-corpus overlap.

\section{\uppercase{Method}}
\label{sec:method}

\subsection{Persistent Artifact, Operations, and Scope}
We model a dialogue as successive state transitions over one authoritative ABC artifact. Let $s_{t-1}$ be the current artifact, $u_t$ the user request, $o_t$ the resolved operation, and $r_t$ its transformation scope. An editing turn can be written as
\begin{equation}
    s_t = F_{o_t}(s_{t-1},u_t,r_t),
    \label{eq:state-transition}
\end{equation}
where $F_{o_t}$ must satisfy the preservation conditions associated with the operation. The resulting $s_t$ becomes the state available to turn $t+1$.

A creation request establishes a new state. An editing request interprets the user instruction relative to the current state, modifies only the material authorized by the operation, preserves its invariants, and returns the revised artifact. In the current implementation, $r_t$ is the whole artifact for \op{Add chords} and \op{Transpose}, and a selected bar span for \op{Inpainting}. Thus, for \op{Add chords}, melody preservation is an invariant; for \op{Inpainting}, material outside the selected bars must remain unchanged; and for \op{Transpose}, structural and rhythmic relations must be preserved while tonal content changes.

\begin{figure*}[t]
\centering
\includegraphics[width=0.94\textwidth]{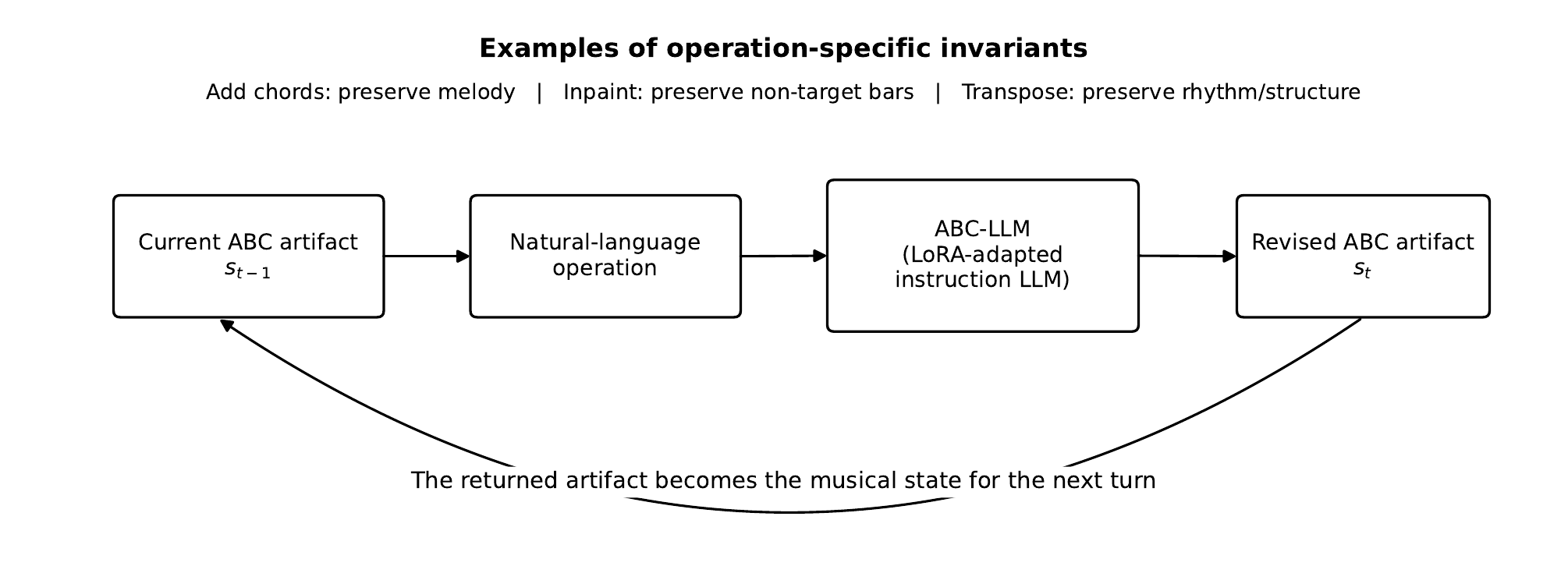}
\caption{Operation-aware state transition over a persistent ABC artifact. The requested operation determines both what may change and what must be preserved; the revised artifact becomes the state for the next turn.}
\label{fig:state-transition}
\end{figure*}

Figure~\ref{fig:state-transition} summarizes this interaction contract. A response must first contain parseable ABC. Only then can an operation-specific checker compare the requested change and the expected invariants. Musical-feature similarity and corpus overlap are evaluated separately rather than folded into the same correctness score.

\subsection{ABC Representation and Musical Domain}
\begin{figure}[t]
{\raggedright\ttfamily\normalsize
X:1\par
L:1/4\par
M:4/4\par
K:C\par
\texttt{"CM7" E F G A | "FM7" A G F E |}\par}
\vspace{1.5mm}
\centering
\includegraphics[width=\linewidth]{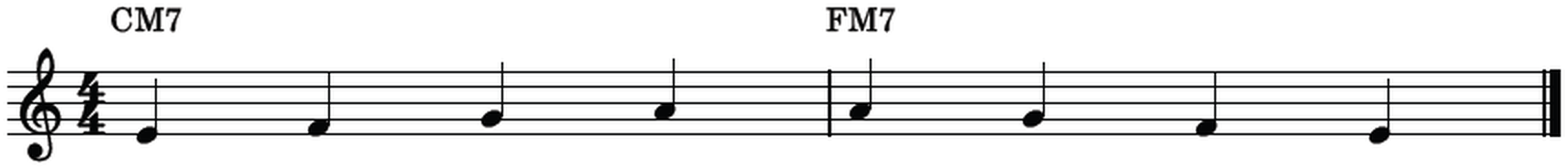}
\caption{A minimal lead-sheet example showing the same musical material represented as ABC text (top) and as conventional notation (bottom).}
\label{fig:abc-leadsheet}
\end{figure}

ABC is plain text, includes headers such as meter and key, and can encode quoted chord symbols together with melody. The same artifact can therefore be passed directly between user and LLM and parsed for comparison across turns. Our experiments use primarily monophonic melody with optional chord annotations, corresponding to a lead-sheet-like setting. Figure~\ref{fig:abc-leadsheet} gives a minimal example of the relation between ABC text and its conventional score rendering.

The empirical domain is the Irish Massive ABC Notation (IrishMan) corpus released with TunesFormer \cite{wu2023tunesformer}. It contains 216,284 tunes, with 214,122 in the reported training split and 2,162 in validation. Dialogue construction uses the training split; held-out material is reserved for evaluation baselines. The study therefore establishes in-domain behavior for Irish traditional music rather than cross-genre generalization.

\subsection{Operation-Aware Dialogue Dataset}
The dataset exposes five user-facing operations: \op{Create a melody}, \op{Create a melody with chords}, \op{Add chords}, \op{Inpainting}, and \op{Transpose}. Creation establishes a new artifact. Chord addition pairs an unannotated melody with its annotated counterpart. Inpainting replaces one to four parsed bars while copying material outside the target. Transposition applies a sampled interval to the current artifact, including chord symbols when present.

Six interaction scenarios combine melody-only and chord-bearing material with two ways to start: asking the assistant to create an artifact or supplying one to edit. A concrete chord-bearing path is, for example, \op{Create a melody} $\rightarrow$ \op{Add chords} $\rightarrow$ \op{Inpainting} (bars 3--5) $\rightarrow$ \op{Transpose}. After every arrow, the assistant output becomes the current artifact for the next request. Other scenarios start directly from user-supplied material or use \op{Create a melody with chords}. Natural-language variation is supplied by 856 template records across 19 internal prompt types.

The constructed artifact contains 496,038 JSONL records; 495,853 contain at least one supervised operation. Across them are 3,241,509 user--assistant operation instances, with a mean of 6.535 operations per record and a range of 1--20 among non-empty records.

\subsection{LoRA Adaptation}
The base checkpoint is \modelname\ \cite{grattafiori2024llama3}. We adapt it by supervised instruction fine-tuning with Low-Rank Adaptation (LoRA), which freezes the base weights and learns low-rank update matrices \cite{hu2022lora}. The configuration uses rank 64, $\alpha=16$, dropout 0.1, no bias adaptation, and causal-language-model task type. Training uses a maximum sequence length of 8,192 tokens for two epochs, per-device batch size one, learning rate $10^{-4}$, \texttt{paged\_adamw\_32bit}, cosine scheduling, weight decay 0.1, warm-up ratio 0.05, maximum gradient norm 1.0, and gradient checkpointing. For the experiments reported here, the trained adapter is merged into the base model; we refer to the result provisionally as \abcllm.

\subsection{Evaluation Protocol}
\label{sec:evaluation-protocol}
Evaluation separates three questions: (1) does the model return usable ABC and perform the requested operation while preserving its invariants? (2) how do selected musical characteristics compare with IrishMan references? and (3) how much sequence overlap occurs with training material relative to held-out data?

\begin{figure*}[t]
\centering
\includegraphics[width=0.94\textwidth]{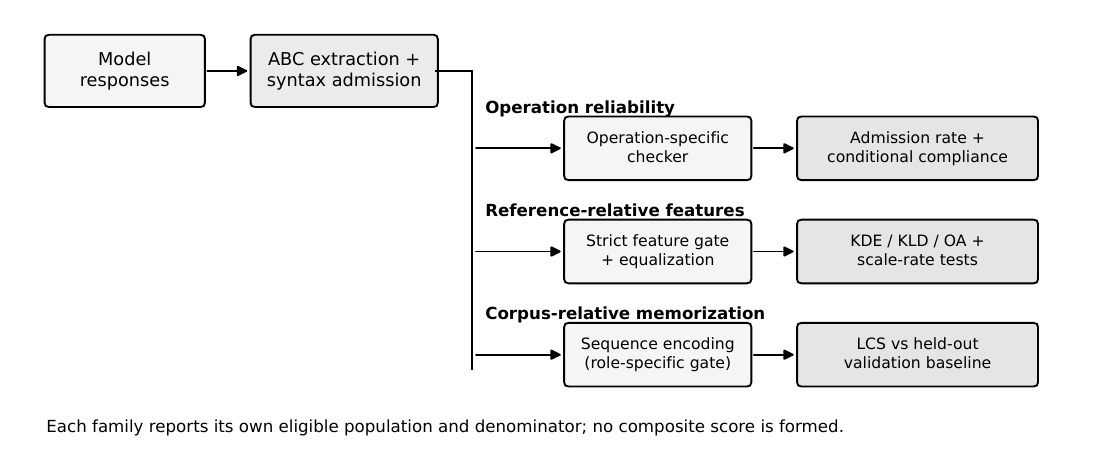}
\caption{Denominator-aware evaluation architecture. Syntax admission is shared, after which operation reliability, reference-relative musical features, and corpus-relative memorization follow separate protocols and populations.}
\label{fig:evaluation-axes}
\end{figure*}

Each model is evaluated on 500 dialogues. Two hundred and fifty form a melody-only chain with three responses: generation, inpainting, and transposition. The other 250 form a chord-bearing chain with four responses: melody generation before harmonization, \op{Add chords}, inpainting of melody and harmony, and transposition of melody and harmony. These are the \emph{seven evaluated output states}, with 250 attempts per state and 1,750 attempts per model. The separate dataset operation \op{Create a melody with chords} is trained but is not a dedicated state in this evaluation protocol. Figure~\ref{fig:evaluation-axes} shows how the three analytical families branch after syntax admission while retaining distinct gates and denominators.

\subsubsection{Admission and Operation Compliance}
Responses first pass ABC extraction and parser-owned syntax admission. An \emph{admitted} response is one that reaches the relevant operation checker; \emph{compliance} asks how well that checker judges the requested transformation. For example, if an \op{Add chords} response is malformed ABC, it is not admitted and cannot receive a conditional compliance score. If it is valid ABC but changes the melody while adding chords, it is admitted but receives poor compliance. Reporting both quantities prevents a high conditional score on a small, selectively successful subset from being mistaken for end-to-end reliability.

\subsubsection{Reference-Relative Musical Characteristics}
\begin{figure}[t]
\centering
\includegraphics[width=0.96\linewidth]{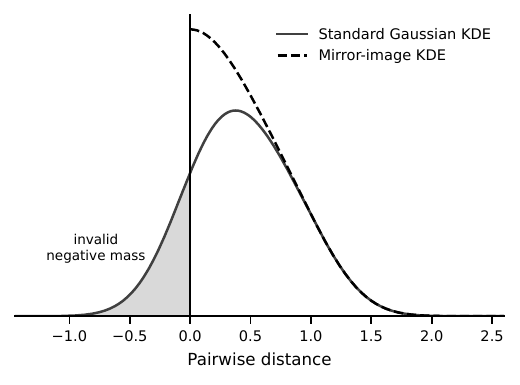}
\caption{Schematic boundary effect for non-negative pairwise distances. Standard Gaussian KDE assigns mass below zero; the Schuster mirror-image estimator reflects that contribution back onto the valid support.}
\label{fig:kde-boundary}
\end{figure}

The musical-feature axis is formative, not an absolute quality score. It adapts the distributional approach of Yang and Lerch \cite{yang2020evaluation}. Nine pitch/rhythm comparisons use pairwise Euclidean distances, Kullback--Leibler divergence (KLD), and overlap area (OA). Because those distances have support on $[0,\infty)$, standard symmetric Gaussian kernel density estimation (KDE) can leak probability mass below zero when observations cluster near the boundary. \textit{Musimetrics} therefore uses Schuster's mirror-image support correction \cite{schuster1985}: each non-negative observation is paired with its reflection across zero before the two distributions are compared (Figure~\ref{fig:kde-boundary}). This is a small methodological contribution relative to applying the Yang--Lerch KDE directly, and is particularly relevant when reference--reference and reference--candidate distances are concentrated near zero. Chord-bearing states additionally use the chord-tone/non-chord-tone ratio (CTnCTR); pitch- and chord-in-scale rates use score-level Mann--Whitney tests with Holm correction.

\subsubsection{Corpus-Relative Memorization}
Possible memorization is characterized using Longest Common Subsequence (LCS) similarity, not treated as a plagiarism detector. Note pitch classes and chord pitch-class sets are encoded relative to the resolved tonic, making the representation invariant to consistent global transposition while retaining durations and event order. Candidate-to-training overlap is compared with held-out validation-to-training overlap. A descriptive flag uses $\mathrm{copy}>0.8$; Mann--Whitney tests compare the distributions. These results concern corpus-relative sequence overlap, not plagiarism, authorship, or intent.

\section{\uppercase{Experimental Results}}
\label{sec:results}

\subsection{Admission and Compliance}
The principal result concerns whether the model can participate reliably in the workflow. Of 1,750 base-model attempts, 1,234 contain extractable ABC and only 514 reach an operation checker; the fine-tuned model yields extractable ABC in all 1,750 attempts and 1,739 reach a checker. Overall checker admission therefore increases from 0.2937 to 0.9937. Conditional operation compliance among admitted turns increases from 0.7205 to 0.9798.

\begin{figure}[t]
\centering
\includegraphics[width=0.98\linewidth]{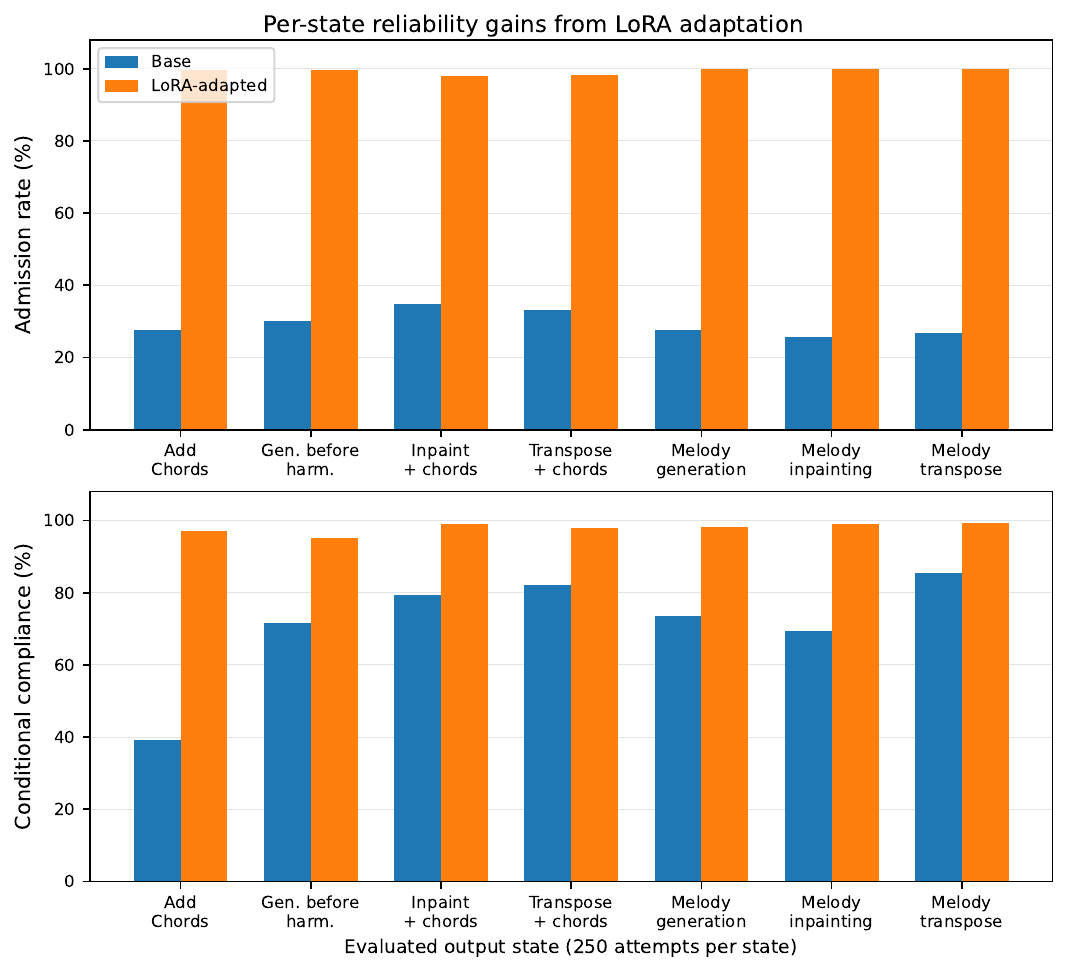}
\caption{Per-state decomposition of Table~\ref{tab:compliance}. Admission and compliance both improve across all seven evaluated output states. Melody transposition is the clearest case where the base model retains some conditional competence once a response reaches the checker, yet still suffers from poor end-to-end admission.}
\label{fig:per-operation-reliability}
\end{figure}

The distinction is important. A malformed transposition response contributes to the low admission rate because no checker can evaluate it. A syntactically valid response that reaches the checker but fails to realize the requested key is admitted yet non-compliant. Table~\ref{tab:compliance} reports these two stages separately, and Figure~\ref{fig:per-operation-reliability} makes the per-state pattern visually explicit.

\begin{table}[t]
\caption{Admission and conditional compliance (250 attempts per output state). FT = fine-tuned.}
\label{tab:compliance}
\centering
\scriptsize
\setlength{\tabcolsep}{2.0pt}
\begin{tabular}{@{}lrrrr@{}}
\toprule
State & Base adm. & FT adm. & Base comp. & FT comp.\\
\midrule
Add Chords & .276 & .996 & .3910 & .9702\\
Gen. before harm. & .300 & .996 & .7148 & .9514\\
Inpaint + chords & .348 & .980 & .7942 & .9904\\
Transpose + chords & .332 & .984 & .8221 & .9796\\
Melody generation & .276 & 1.000 & .7346 & .9832\\
Melody inpainting & .256 & 1.000 & .6944 & .9915\\
Melody transpose & .268 & 1.000 & .8548 & .9923\\
\midrule
Overall & .2937 & .9937 & .7205 & .9798\\
\bottomrule
\end{tabular}
\end{table}

\begin{figure}[t]
\centering
\includegraphics[width=0.98\linewidth]{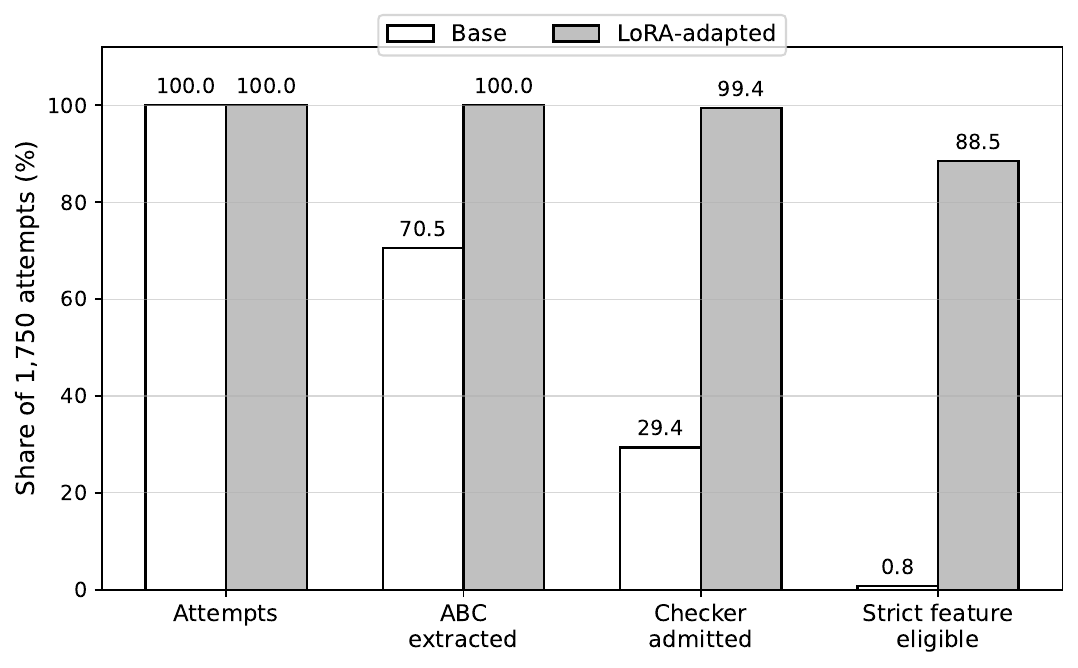}
\caption{Population retained at successive evaluation stages, expressed as a percentage of the 1,750 attempts per model. ``Strict feature eligible'' is the pre-equalization musical-feature population.}
\label{fig:population-funnel}
\end{figure}

Figure~\ref{fig:population-funnel} visualizes the same end-to-end population effect. The base model loses most attempts before operation checking and almost all before strict feature analysis, whereas the adapted model retains nearly the full response population through both stages.

Operation-level diagnostics show that the gain extends beyond syntax. In the fine-tuned response set, chord addition preserves 4,639 of 4,644 compared melody bars and adds the expected quoted chord symbols in 224 of 249 admitted turns. Inpainting preserves 8,886 of 9,043 compared non-target bars and modifies 1,168 of 1,230 target bars. Transposition matches the requested key in 492 of 496 admitted turns. The adaptation therefore improves both requested changes and operation-specific preservation.

\subsection{Reference-Relative Musical Characteristics}
The greatly enlarged admissible population also changes what can be characterized musically. Only 14 of 1,750 base-model outputs pass the strict musical-feature gate. For the fine-tuned model, 1,548 pass before scenario equalization and 1,471 are used afterwards. Consequently, all 63 planned Yang--Lerch comparisons (nine features across the seven states defined in Section~\ref{sec:evaluation-protocol}) are calculable for the adapted model; the base population is too sparse for an equivalent characterization.

For the fine-tuned model, overlap area (OA) values across the nine features range from 0.8952--0.9724 for Add Chords, 0.8297--0.9687 for generation before harmonization, 0.9315--0.9853 for inpainting with chords, 0.8301--0.9862 for transposition with chords, 0.8727--0.9711 for melody generation, 0.8875--0.9731 for melody inpainting, and 0.7972--0.9730 for melody transposition. These values describe distributional agreement for selected features, not absolute aesthetic quality.

CTnCTR is calculable for all three chord-editing states, with OA 0.9718 for Add Chords, 0.9652 for inpainting with chords, and 0.9696 for transposition with chords. Pitch- and chord-in-scale comparisons are ceiling-heavy: reference and candidate medians are 1.0 in all ten comparisons, while candidate rates are slightly higher under the implemented definitions. These measurements should therefore be read as corpus-relative descriptors.

\subsection{Corpus-Relative Memorization}
For the adapted model, all seven rates above the strict $\mathrm{copy}>0.8$ LCS threshold are below their corresponding held-out validation rates: 15.60\% for chord-bearing references and 27.60\% for melody-only references. Add Chords differs from its baseline distribution at the unadjusted level ($p=0.0072$), yet its flagged rate is 14.47\%, below the 15.60\% held-out rate; the other six comparisons have $p$ values from 0.0622 to 0.5852. We therefore find no systematic increase in high-overlap sequences relative to the held-out baselines under this representation and threshold. This does not prove absence of memorization or plagiarism.

\subsection{Concrete Multi-Turn Example}
\label{sec:persistent-example}
\begin{figure*}[t]
\centering
\includegraphics[width=0.98\textwidth]{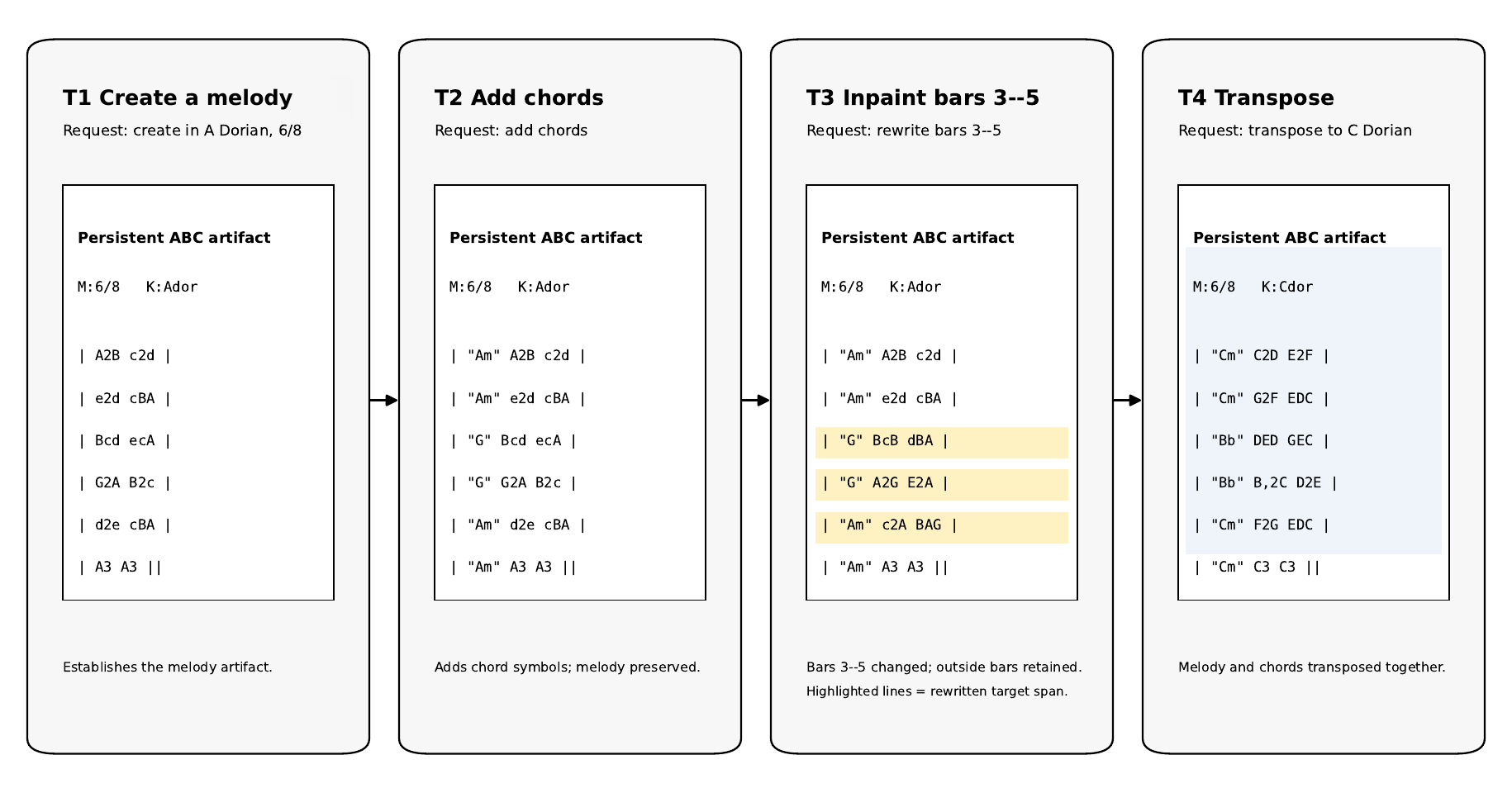}
\caption{Illustrative rendering of the four-turn persistent-artifact workflow summarized in Table~\ref{tab:dialogue-example}. Each turn produces a revised ABC artifact that becomes the direct input to the next request. The example is schematic, but it matches the evaluated semantics: creation establishes the artifact, chord addition preserves melody, inpainting rewrites only the targeted region, and transposition updates both melody and chord symbols together.}
\label{fig:persistent-dialogue}
\end{figure*}

An author-selected evaluated dialogue (song\_id=44) makes the state semantics concrete. Table~\ref{tab:dialogue-example} summarizes how each response becomes the input artifact for the next turn, while Figure~\ref{fig:persistent-dialogue} gives a compact visual rendering of the same persistent-artifact logic.

\begin{table}[t]
\caption{A four-turn evaluated dialogue showing persistent state transitions.}
\label{tab:dialogue-example}
\centering
\scriptsize
\setlength{\tabcolsep}{3pt}
\begin{tabular}{@{}p{0.10\linewidth}p{0.31\linewidth}p{0.49\linewidth}@{}}
\toprule
Turn & Request & Observable state effect\\
\midrule
1 & Create in A Dorian, 6/8 & Establishes the melody artifact.\\
2 & Add chords & Inserts \texttt{"Am"} and \texttt{"G"}; melody unchanged.\\
3 & Inpaint bars 3--5 & Rewrites target region; other bars retained.\\
4 & Transpose to C Dorian & Transposes melody and chord symbols.\\
\bottomrule
\end{tabular}
\end{table}

Across the fixed response sets, LoRA adaptation increases ABC extraction (1,234 to 1,750), checker admission (514 to 1,739), strict feature eligibility (14 to 1,548 before equalization), and conditional compliance (0.7205 to 0.9798). The main empirical conclusion is therefore operational: adaptation makes the tested interaction contract reliable in this domain, while the musical-feature and LCS axes answer separate questions about the resulting admitted population.

\section{\uppercase{Prototype Integration with NONOTO}}
\label{sec:prototype}

To make the intended use of \abcllm\ concrete, we developed an experimental integration with an extended version of NONOTO, the model-agnostic web interface introduced by Bazin and Hadjeres for interactive composition by inpainting \cite{bazin2019nonoto}. NONOTO already provides a score-oriented interaction environment with audio playback and real-time MIDI output; our extension adds the natural-language/ABC path needed for the operation-aware workflow. A user can inspect or edit musical material, request a compositional operation in natural language, receive revised ABC, visualize the resulting notation, and audition the result through the interface's existing rendering/output facilities.

\begin{figure}[t]
\centering
\includegraphics[width=\linewidth]{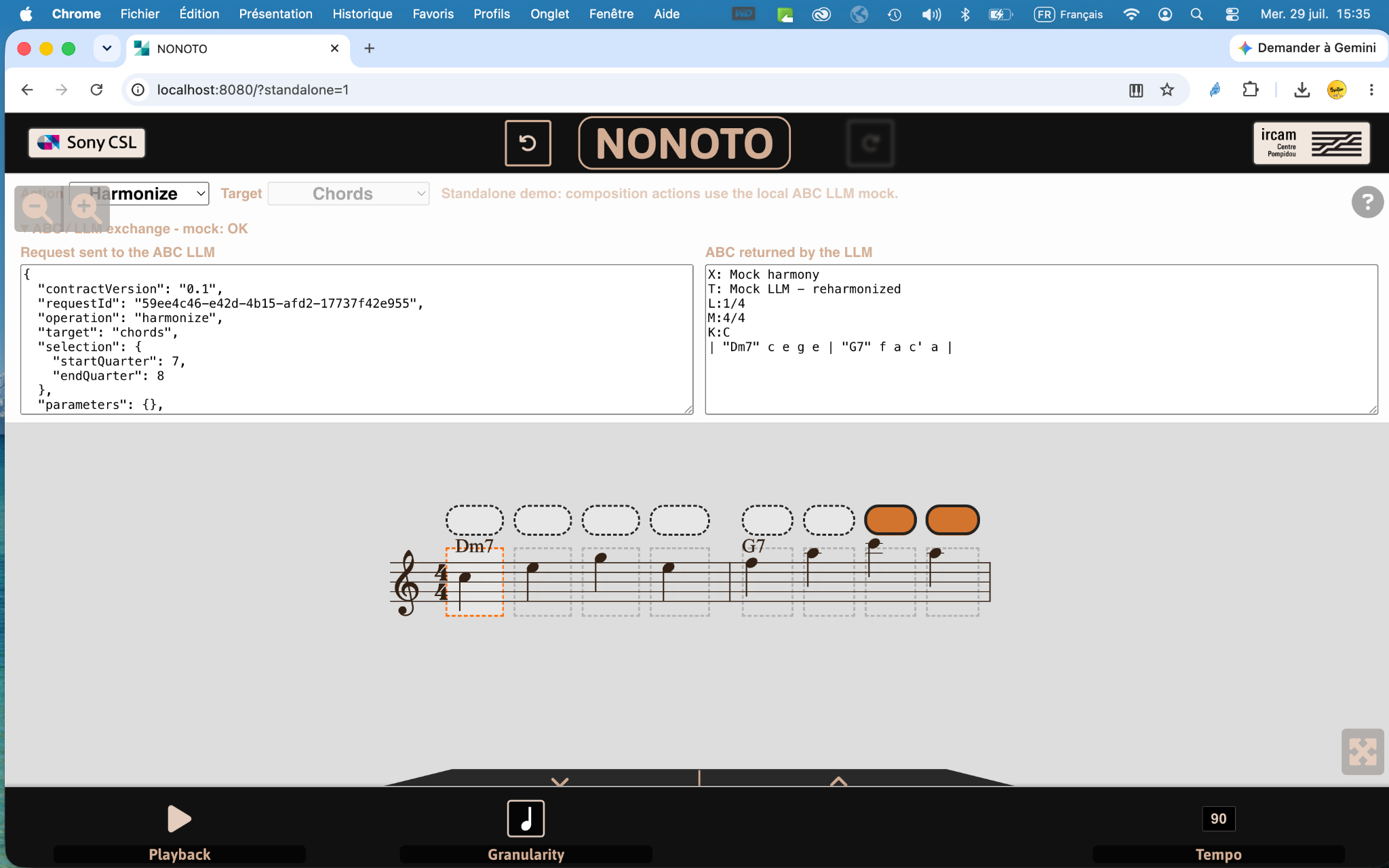}
\caption{Prototype of the extended NONOTO score editor illustrating the natural-language/ABC interaction workflow considered in this work.}
\label{fig:nonoto}
\end{figure}

The prototype is not presented as a professional composition environment or as an HCI evaluation. Its role is to demonstrate the architectural path from a natural-language request to an updated, inspectable, and playable symbolic artifact. Figure~\ref{fig:nonoto} illustrates the workflow.

\section{\uppercase{Discussion}}
\label{sec:discussion}

\paragraph{Scope of the claim.}
The central contribution is not the observation that task-specific fine-tuning improves a model. LoRA is the enabling mechanism, and the base/adapted comparison is a feasibility test of a more specific hypothesis: whether a general-purpose instruction LLM can learn to sustain an operation-aware interaction contract over one evolving musical artifact. The strongest conclusion supported by the experiments is therefore narrower than ``fine-tuning improves music generation.'' The unadapted instruction model shows conditional competence on some admitted operations---transposition is the clearest example---but is unreliable end-to-end because many responses fail extraction, parsing, or checker admission. After adaptation, almost every response reaches an operation checker and admitted responses satisfy the requested contract at a high rate.

This result complements rather than supersedes recent editing systems. BeatEdit builds explicit symbolic edit mechanisms, while MusiChat maintains persistent conversational refinement through a hybrid LLM/symbolic architecture \cite{gu2026beatedit,liao2026musichat}. Our experiment supports a different design point: operation semantics, persistence, and preservation behavior can be learned directly by a general-purpose instruction model under the tested supervision and domain. It does not establish that this architecture is preferable to dedicated editors, hybrid systems, deterministic algorithms, or constraint solvers.

This interpretation also clarifies what supervision teaches. Editing examples expose before/after relations: some parts of the artifact are authorized to change and others are invariants. The gains in chord insertion, preservation outside inpainting targets, target modification, and transposition provide evidence of operation-aware behavior under these scenarios; they are not evidence of abstract musical understanding.

The musical-feature results answer a different question. The adapted model yields enough valid material for reference-relative characterization across the seven output states listed in Section~\ref{sec:evaluation-protocol}. The selected distributions often substantially overlap IrishMan references, but these metrics are formative descriptors, not a scalar definition of musical quality. Human listening and musician-centered studies remain necessary for coherence, usefulness, playability, stylistic plausibility, perceived control, and the practical value of successive edits.

The memorization analysis is similarly bounded. Tonic-relative, chord-aware LCS provides a reproducible sequence-overlap measure and a held-out baseline. The results do not reveal a systematic excess of thresholded overlap, but neither low overlap nor a non-significant test establishes absence of memorization, plagiarism, authorship, or intent.

Transposition also illustrates a boundary of the architecture. An exact symbolic transposition algorithm is preferable when that isolated operation is all that is required. Here, transposition is diagnostically useful because it tests whether heterogeneous operations can be requested through one language interface and applied to the same persistent artifact. More generally, an LLM assistant need not replace deterministic algorithms or solvers when exactness is available.

Finally, the scope is restricted to Irish traditional music, five operations, curated prompt templates, and a primarily monophonic lead-sheet representation. The current operation set also mixes fixed scopes: chord addition and transposition are whole-artifact transformations, while inpainting is local. The study establishes model-side reliability for this interaction contract under those choices, not cross-genre generalization, superior musical quality, or human--AI co-creativity.

\section{\uppercase{Future Work}}
\label{sec:future}

A first extension is to decouple the requested operation from its transformation scope. The current prototype uses a whole-artifact scope for \op{Add chords} and \op{Transpose}, and selected bars for \op{Inpainting}. A more general composer's assistant should support combinations such as ``transpose bars 5--8'', ``harmonize the chorus'', or ``reharmonize measures 9--12'', with preservation requirements derived from both the operation and the selected region.

A second extension concerns musical domains. Ongoing work is evaluating the approach beyond Irish traditional music, with jazz and bossa nova as initial targets, including work within an ongoing undergraduate final project. LoRA suggests a practical architecture for specialization: the base weights can remain fixed while small domain adapters are learned offline and selected at runtime \cite{hu2022lora}. This resembles, at the interaction level, selecting a corpus-specific context in systems such as FlowComposer while using a different learning mechanism.

Musician-centered evaluation is the next validation layer rather than evidence claimed by the present paper. The NONOTO integration provides a basis for studies of usefulness of successive edits, preservation of musical intent, perceived control, workflow compatibility, and authorship/ownership perceptions. Where task definitions and outputs are comparable, future experiments can also apply common operation-level measures to systems such as BeatEdit and MusiChat rather than relying only on within-model base/adapted comparison.

A deeper research direction is to combine LLM generation with explicit Constraint Programming (CP). GenCP formulates constrained generation as a Constraint Satisfaction Problem and combines language-model predictions with constraint propagation \cite{bonlarron2025gencp}; a recent survey reviews this broader CP/generative-model lineage \cite{bonlarron2026survey}. For ABC, such a hybrid may depend strongly on tokenization. MidiTok demonstrates the value of musically designed tokenization for MIDI \cite{fradet2021miditok}; an analogous ABC-oriented tokenizer could align variables with note events, durations, chord symbols, and barlines. The LLM could then interpret intent and propose musically plausible material while a solver guarantees selected hard properties.

Further directions include polyphony, richer harmony, additional operations, longer and less templated dialogues, recovery from invalid intermediate states, and explicit study of cultural/provenance issues when adapting to additional repertoires.

\section{\uppercase{Conclusion}}
\label{sec:conclusion}

This paper reframed conversational symbolic music generation as a problem of incremental composition over a persistent artifact. The question is not simply whether an LLM can generate music, but whether a general-purpose instruction model can reliably interpret successive compositional requests, change the material authorized by each operation, preserve the required invariants, and carry the resulting score into the next turn.

Under 1,750 attempted output states per model, LoRA adaptation of Llama 3.1 8B Instruct raises checker admission from 29.37\% to 99.37\% and conditional operation compliance from 0.7205 to 0.9798. It also raises strict musical-feature eligibility from 14 to 1,548 outputs before equalization. Within the tested IrishMan domain, the adapted model therefore becomes a reliable participant in the specified interaction contract. The result is a feasibility finding about model-side interaction reliability, not a claim that fine-tuning itself is novel or that the resulting music is aesthetically superior.

The complementary analyses deliberately support weaker claims. Distributional metrics describe selected musical properties relative to IrishMan rather than absolute musical quality, and tonic-relative LCS characterizes corpus overlap rather than proving or disproving plagiarism. The NONOTO integration illustrates how the learned behavior can be embedded in an edit--render--listen--revise workflow. Future work will generalize operation scope, test domain-specific adapters and musician-centered interaction, and investigate structured ABC tokenization with explicit constraint solving.

\section*{Acknowledgements}
The authors thank Gaëtan Hadjeres for suggesting, at the inception of this project, that we investigate whether a general-purpose large language model could support an incremental and interactive music-composition workflow. Jean-Pierre Briot acknowledges FAPERJ (Brazil) for partial support of this study through a PV research fellowship.

\section*{Generative AI Use Disclosure}
The research problem and approach, system design and implementations, experimental protocol and execution, scientific analysis, and an initial complete presentation of the work were developed by the authors without generative-AI assistance. ChatGPT (OpenAI) was subsequently used for English-language editing and to help refine the organization, exposition, and rhetorical clarity of the Introduction, Related Work, Results, Discussion, Future Work, and Conclusion. The authors reviewed and edited all AI-assisted text, verified the cited sources and scientific claims, and take full responsibility for the final content \cite{openai2026chatgpt}.

\bibliographystyle{apalike}
\bibliography{references}
\end{document}